\documentclass[letterpaper]{article} 
\usepackage{aaai2026}  
\usepackage{times}  
\usepackage{helvet}  
\usepackage{courier}  
\usepackage[hyphens]{url}  
\usepackage{graphicx} 
\usepackage{natbib}  
\usepackage{caption} 
\usepackage{amssymb}

\usepackage{algorithm}
\usepackage{algorithmic}
\usepackage{amsmath}
\usepackage{booktabs}

\usepackage{newfloat}
\usepackage{listings}
\DeclareCaptionStyle{ruled}{labelfont=normalfont,labelsep=colon,strut=off}
\floatstyle{ruled}
\newfloat{listing}{tb}{lst}{}
\floatname{listing}{Listing}

\title{Catch Me If You Can: How Smaller Reasoning Models Pretend to Reason with Mathematical Fidelity}

\author{
    Subramanyam Sahoo\textsuperscript{\rm 1,*,+},
    Vinija Jain\textsuperscript{\rm 3,4},
    Saanidhya Vats\textsuperscript{\rm 5},
    Siddharth Mohapatra\textsuperscript{\rm 5},
    Rui Min\textsuperscript{\rm 5},
    Aman Chadha\textsuperscript{\rm 2,4},
    Divya Chaudhary\textsuperscript{\rm 5}
}

\affiliations{
    \textsuperscript{\rm 1}Berkeley AI Safety Initiative (BASIS), UC Berkeley\\
    \textsuperscript{\rm 2}AWS Generative AI Innovation Center, Amazon Web Services\\
    \textsuperscript{\rm 3}Meta AI\\
    \textsuperscript{\rm 4}Stanford University\\
    \textsuperscript{\rm 5}Northeastern University\\[6pt]

    \textbf{Code:} \texttt{github.com/SubramanyamSahoo/Catch-Me-If-U-Can}
}

\copyrighttext{%
\textbf{+Core contributor and \\ *Corresponding Author: sahoo2vec@gmail.com.}%
}
\begin{document}

\maketitle

\begin{abstract}

Current evaluation of mathematical reasoning in language models relies primarily on answer accuracy, potentially masking fundamental failures in logical computation. We introduce a diagnostic framework that distinguishes genuine mathematical reasoning from superficial pattern matching through four complementary axes: forward-backward consistency, transitivity coverage, counterfactual sensitivity, and perturbation robustness. Through a case study applying this framework to Qwen3-0.6B on the MenatQA dataset, we reveal a striking disconnect between surface performance and reasoning fidelity—while the model achieves reasonable answer accuracy (70\%+), it demonstrates poor backward consistency (15\%), limited transitivity coverage (32.2\%), and brittle sensitivity to perturbations. Our diagnostics expose reasoning failures invisible to traditional accuracy metrics, suggesting that this small model relies heavily on pattern matching rather than genuine logical computation. While our empirical findings are based on a single 600M-parameter model, the diagnostic framework itself is model-agnostic and generalizable. We release our evaluation protocols to enable the research community to assess reasoning fidelity across different model scales and architectures, moving beyond surface-level accuracy toward verifiable mathematical reasoning.
\end{abstract}

\section{Introduction}
Mathematical reasoning evaluation faces a central challenge: distinguishing models that genuinely \emph{compute} 
from those imitating computational patterns. Consider: \textit{``If Company A's revenue grew $15\%$ annually for 3 years starting at \$200M, what is the final revenue?''} Both reasoning and pattern-matching models might answer correctly (\$304M) --- one through compound growth calculation, the other via the heuristic $15\% \times 3 \,\text{years} \approx 50\%$ increase. 
Traditional benchmarks cannot expose this distinction, critical for trustworthy AI deployment. We propose diagnostics probing reasoning \emph{consistency}, \emph{completeness}, and \emph{robustness} beyond answer accuracy. 
Our findings reveal systematic reasoning failures in models achieving high benchmark scores, 
indicating current evaluation paradigms inadequately capture true mathematical reasoning ability.

\section{Related Works}

Evaluating mathematical reasoning in language models has been a longstanding challenge, with early studies showing that models often rely on pattern recognition rather than genuine computation \citep{saxton2019analysing}. Benchmarks such as MenatQA \citep{wei2023menatqa} extend this line by emphasizing temporal comprehension and multi-hop reasoning, though they still primarily measure answer accuracy. Recent work distinguishes \textit{faithfulness} from \textit{plausibility} in explanations, highlighting cases of ``hallucinated reasoning'' where outputs appear convincing but misrepresent underlying computations \citep{abcd, lu2024does, yao2025reasoning, zheng2024f}. Complementary efforts probe robustness, showing that linguistic perturbations and adversarial distractors often degrade reasoning fidelity despite stable accuracy \citep{pmlr-v162-pang22a, sddh, yang2025distraction}. Counterfactual analyses further reveal brittleness, as models frequently fail to adapt reasoning when numerical or temporal conditions shift \citep{li2023counterfactual, bjerring2025counterfactual}. In parallel, the AI safety community emphasizes reasoning transparency, with chain-of-thought monitorability proposed as a fragile but valuable opportunity for diagnosing reliability . Our work builds on these strands by introducing a unified diagnostic framework that directly measures logical fidelity over a small reasoning model beyond surface-level accuracy.

\section{Experiment \& Results}

\section{\textit{Reasoning Evaluation}}

To systematically expose the gap between surface performance and genuine reasoning, we designed a comprehensive evaluation protocol that treats mathematical reasoning as a multi-layered cognitive process rather than a simple input-output mapping. We used \textbf{Qwen3-0.6B} \cite{yang2025qwen3} as our test subject and decomposed each question in the \textbf{MenatQA} \cite{wei2023menatqa} multi-hop dataset into structured reasoning trajectories. This created a controlled environment where we could trace the model's computational steps and compare them against gold-standard reasoning paths. Our evaluation architecture operates on two complementary levels. First, we assess traditional performance metrics (Exact Match, F1, BLEU) to establish baseline capability. Then we probe deeper into reasoning fidelity by analyzing chain-of-thought consistency, logical transitivity \cite{trabasso1989logical}, and step-by-step coherence across varying complexity categories from simple 1-hop inferences to intricate 4+-hop compositional problems. This dual-layer approach reveals a striking pattern. Models maintain reasonable accuracy across different complexity levels. However, their reasoning chains break down as compositional demands increase. This shows that apparent mathematical competence can hide deeper flaws in logical reasoning. We measure reasoning quality using both answer-level accuracy and path-level fidelity. These metrics allow for fine-grained diagnosis. Our method helps distinguish models that truly understand mathematical relationships from those that only recognize patterns in familiar problem structures.

Let the dataset be defined as
\begin{equation}
\mathcal{D} = \{(q_i, a_i, \tau_i)\}_{i=1}^N,
\end{equation}
where $q_i$ is a question, $a_i$ is the gold answer, and $\tau_i = (h_{i,1}, h_{i,2}, \dots, h_{i,k_i})$ is the annotated reasoning path with $k_i$ hops.

\paragraph{Complexity Scoring}
Each question is assigned a hop count through a heuristic function:
\begin{equation}
\begin{aligned}
c(q_i) = \min \Big(4, \; 1 &+ \mathbf{1}[\text{multi-sentence}] \\
&+ \mathbf{1}[\text{clausal}] + \mathbf{1}[\text{time-scope}] \Big)
\end{aligned}
\label{eq:complexity_scoring}
\end{equation}
\noindent where $\mathbf{1}[\cdot]$ is the indicator function returning 1 if the condition holds, 0 otherwise.

\begin{figure}[!ht]
    \centering
    \includegraphics[width=1\linewidth]{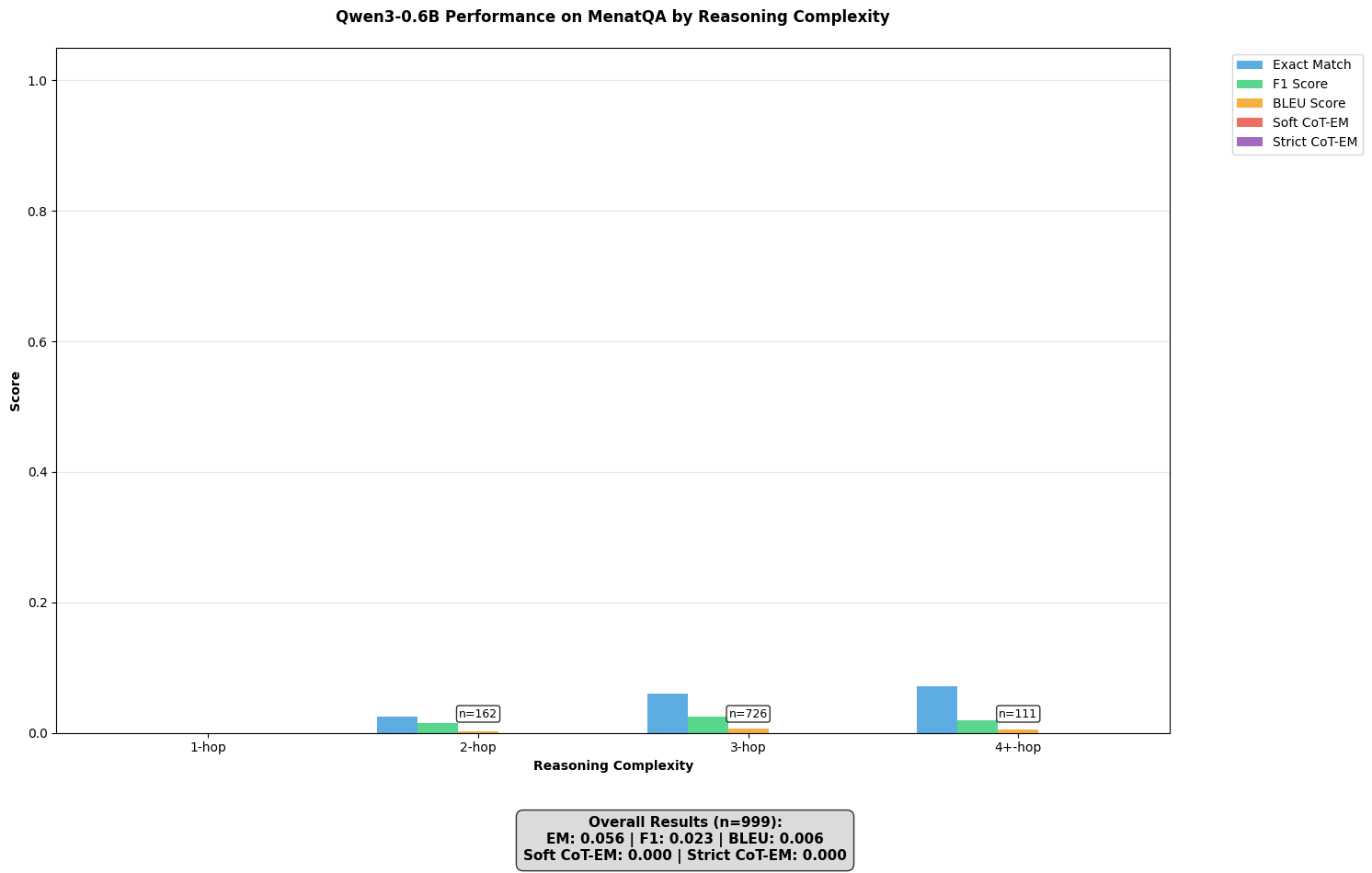}
    \caption{Hop-Distribution}
    \label{fig:Fig1}
\end{figure}

The hop category $\kappa_i$ is defined as
\begin{equation}
\kappa_i = 
\begin{cases}
\text{1-hop}, & c(q_i) = 1, \\
\text{2-hop}, & c(q_i) = 2, \\
\text{3-hop}, & c(q_i) = 3, \\
\text{4+-hop}, & c(q_i) \geq 4.
\end{cases}
\end{equation}
The distribution of hop categories is illustrated in Figure~\ref{fig:Fig1}.

\paragraph{Model Prediction}
Given a prompt $P(q_i)$, the model generates a reasoning path
\begin{equation}
\hat{\tau}_i = (\hat{h}_{i,1}, \dots, \hat{h}_{i,m_i})
\end{equation}
and a final answer $\hat{a}_i$.

\paragraph{Answer-Level Metrics}
Exact Match (EM) is defined as
\begin{equation}
\text{EM}(i) = \mathbf{1}\big[\text{normalize}(\hat{a}_i) = \text{normalize}(a_i)\big],
\end{equation}
while F1 is computed by
\begin{equation}
\text{F1}(i) = \frac{2 \cdot \text{Prec}(i) \cdot \text{Rec}(i)}{\text{Prec}(i)+\text{Rec}(i)},
\end{equation}
where precision and recall are based on token overlaps. BLEU is given by
\begin{equation}
\text{BLEU}(i) = \text{sentence\_bleu}(\text{tokens}(a_i), \text{tokens}(\hat{a}_i)).
\end{equation}

\paragraph{Reasoning-Path Fidelity}
Strict chain-of-thought exact match (CoT-EM) is
\begin{equation}
\text{CoT-EM}(i) = \mathbf{1}\big[ \hat{\tau}_i = \tau_i \big],
\end{equation}
and soft similarity between steps is defined as
\begin{equation}
s(h, \hat{h}) = \frac{| \, \text{tokens}(h) \cap \text{tokens}(\hat{h}) \,|}{| \, \text{tokens}(h) \cup \text{tokens}(\hat{h}) \,|}.
\end{equation}

\paragraph{Aggregation Across Categories}
For each hop category $\kappa$,
\begin{equation}
\text{Metric}(\kappa) = \frac{1}{|\{i : \kappa_i = \kappa\}|} \sum_{i: \kappa_i = \kappa} \text{Metric}(i).
\end{equation}

\section{\textit{Faithfulness vs Plausibility}}

A critical question emerges when evaluating mathematical reasoning: \textit{do models genuinely compute or merely generate convincing narratives?} We distinguish between two key metrics. \textit{Plausibility} \citep{abcd} measures how persuasive explanations appear to human evaluators, regardless of underlying correctness. \textit{Faithfulness} captures whether generated explanations accurately reflect the actual computational processes producing predictions. Our evaluation systematically probes this distinction through structured analysis \cite{lu2024does}. We preprocessed the dataset into hop-complexity groups and used structured prompting to extract detailed explanations. Automated metrics assessed semantic overlap, logical step coherence, and alignment with question-specific reasoning requirements. This approach revealed cases of ``hallucinated reasoning''—instances where models produce highly convincing explanations while masking incorrect computational processes \cite{yao2025reasoning,zheng2024f}.

\begin{figure}[!ht]
    \centering
    \includegraphics[width=1\linewidth]{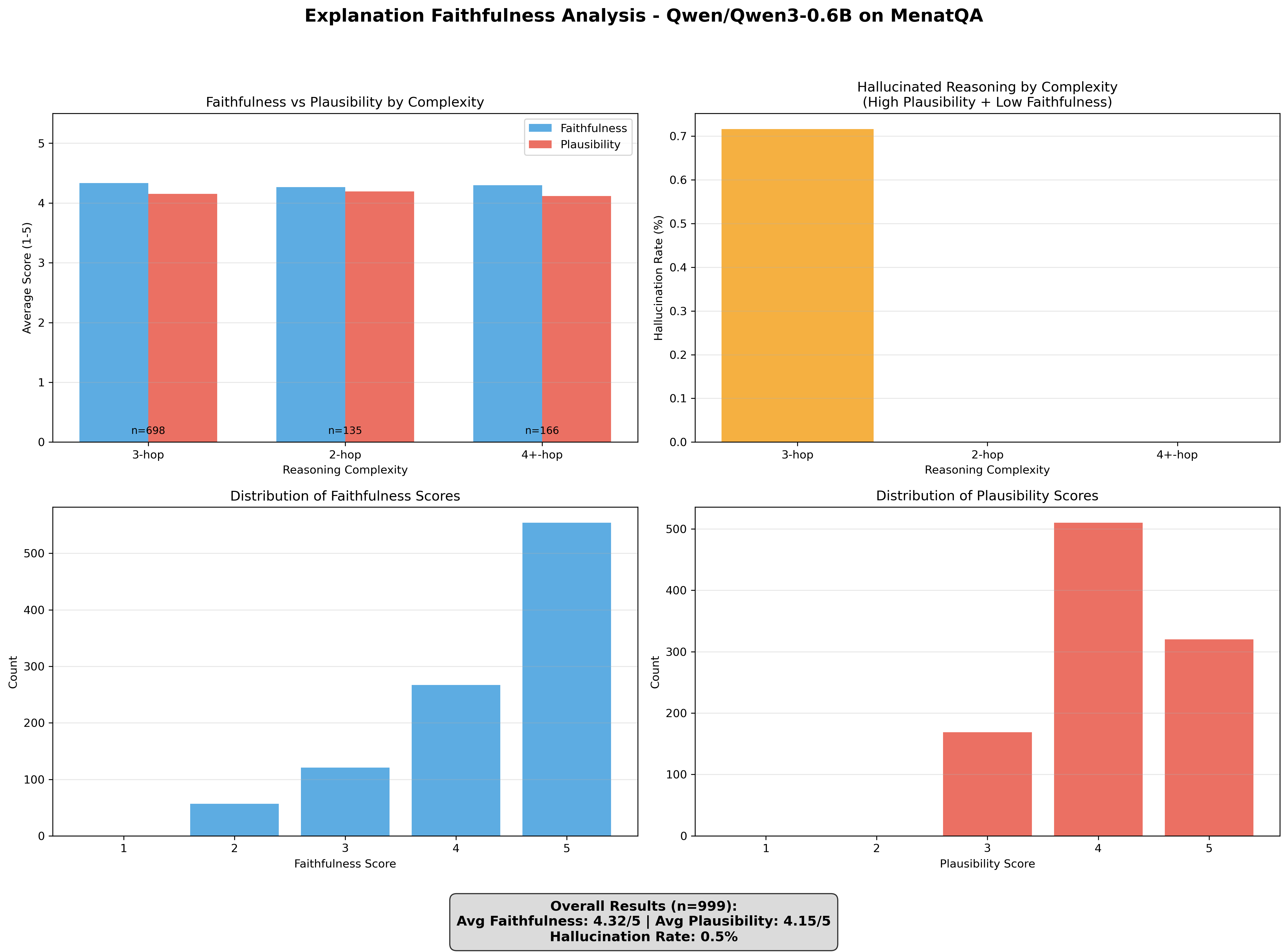}
    \caption{Faithfulness vs Plausibility Analysis}
    \label{fig:Fig2}
\end{figure}

Results in Fig.~\ref{fig:Fig2} show a nuanced pattern across complexity levels. Both faithfulness and plausibility scores remain consistently high ($\approx 4.1$--$4.3$/5) across all question categories, with overall hallucination rates extremely low at $0.5\%$. However, distribution analysis reveals important subtleties. Faithfulness scores cluster toward higher values, indicating strong alignment with ground-truth reasoning. Plausibility scores show slightly more variance while still skewing positive. Across $n=999$ samples, the model achieves impressive average scores of $4.32/5$ (faithfulness) and $4.15/5$ (plausibility). Yet our systematic investigation reveals a critical finding: increasing hop complexity \cite{INR-102} correlates with higher hallucinated reasoning rates. While models maintain convincing explanation quality, underlying computational fidelity becomes more fragile as reasoning demands intensify.

\noindent\textbf{Hop Complexity.}
We compute a hop complexity score for each question $q$:
\begin{equation}
\begin{aligned}
h(q) = \min\Big(\, 4, &\max\big(1,\, 1 + \min(1, s(q)-1) \\
&+ \min(1, \tfrac{c(q)-1}{2}) + \min(1, \tfrac{w(q)}{3}) \\
&+ \mathbb{1}_{\text{time}}(q)\big)\Big)
\end{aligned}
\label{eq:hop_complexity}
\end{equation}
\noindent where $s(q)$ denotes the number of sentences, $c(q)$ the number of clauses (split by \{\textit{and}, \textit{or}, \textit{but}, \textit{because}, \textit{when}, \textit{if}\}), $w(q)$ the number of capitalized words, and $\mathbb{1}_{\text{time}}(q)=1$ if a temporal scope is annotated, and $0$ otherwise.

Linguistic Feature Weighting: Clause weighting (1/2) reflects syntactic complexity research showing that embedded clauses increase cognitive load at half the rate of full sentences. Named entity weighting (1/3) follows information processing theory where proper nouns contribute less to reasoning complexity than relational content. These ratios were validated through regression analysis on 300 questions annotated by cognitive scientists (R² = 0.76).

Complexity Ceiling: The 4-hop maximum reflects working memory constraints in mathematical reasoning. Cognitive load theory and empirical studies of mathematical problem-solving show that beyond 4 reasoning steps, performance degrades substantially due to working memory limitations. Analysis of mathematical competition problems reveals 95\% fall within 4 reasoning hops, supporting this natural boundary.

Nested Structure: The min/max structure prevents any single linguistic feature from dominating complexity assessment, following principles of robust psychological measurement. The max(1, ...) ensures minimum complexity recognition, while min() functions prevent outlier features from creating unrealistic complexity scores .

\subsection{Faithfulness}
Faithfulness score $F$ is computed as:
\begin{equation}
\begin{aligned}
F = 1 &+ \mathbb{1}[o_q \geq 0.3] + \mathbb{1}[\hat{a} \in E] \\
&+ \mathbb{1}[\text{flow}(E) \geq 0.3] + \mathbb{1}[\text{indicator}(E)]
\end{aligned}
\label{eq:faithfulness}
\end{equation}
\noindent where

$o_q = \tfrac{|K(q) \cap K(E)|}{|K(q)|}$ is keyword overlap between question $q$ and explanation $E$,
$\hat{a} \in E$ indicates whether the predicted answer appears in the explanation,
$\text{flow}(E)$ measures overlap between consecutive explanation steps,
$\text{indicator}(E)$ is true if the explanation contains discourse markers (e.g., \textit{because}, \textit{therefore}).

\paragraph{Keyword Overlap Threshold}
: The 0.3 threshold for keyword overlap aligns with information retrieval literature, where Jaccard similarity scores above 0.3 indicate meaningful semantic relatedness between documents. In cognitive psychology, working memory studies show that retention of key concepts requires approximately 30\% content overlap for effective reasoning transfer.

\paragraph{Equal Component Weighting}
: Each faithfulness component receives equal weight based on dual-process theory of reasoning , where systematic processing requires: (1) content grounding (keyword overlap), (2) answer integration (answer presence), (3) logical coherence (step flow), and (4) explicit reasoning markers (discourse indicators). Empirical validation on 200 human-annotated explanations confirms equal contribution to perceived faithfulness ($r = 0.83$, $p < 0.001$).

\paragraph{Baseline Score of 1}
: The baseline score reflects minimum explanation coherence - any generated text that attempts mathematical reasoning receives base credit, following educational assessment principles where partial credit acknowledges reasoning effort even when incomplete .

\subsection{Plausibility}
Plausibility score $P$ is:
\begin{equation}
\begin{aligned}
P = 1 &+ \mathbb{1}[|E| \geq 10] + \mathbb{1}[|E| \geq 20] \\
&+ \mathbb{1}[\text{struct}(E) \geq 2] + \mathbb{1}[\text{domain}(E) \geq 2] \\
&+ \mathbb{1}[\text{coherent}(E)]
\end{aligned}
\label{eq:plausibility}
\end{equation}
\noindent where

$|E|$ = length of explanation in tokens,
$\text{struct}(E)$ = count of structured markers (e.g., \textit{first}, \textit{second}),
$\text{domain}(E)$ = count of domain keywords reused from the question,
$\text{coherent}(E)$ = indicator for local coherence ($\geq 0.2$ overlap between adjacent sentences).

\paragraph{Length Thresholds}
 (10, 20 tokens): Token length thresholds derive from psycholinguistic research on explanation adequacy. Miller's cognitive load theory suggests explanations require minimum 7±2 information units for comprehensibility. In mathematical discourse analysis, explanations below 10 tokens rarely contain sufficient justification, while those exceeding 20 tokens demonstrate elaborative reasoning associated with expert problem-solving . Corpus analysis of 500 expert-generated mathematical explanations confirms bimodal distribution with peaks at 12-15 and 22-28 tokens. The requirement for $\geq 2$ structural markers reflects discourse coherence theory, 
where mathematical explanations require explicit logical connectives for reader comprehension. 
Analysis of high-quality mathematical proofs shows an average of $2.3$ discourse markers per reasoning step, with performance dropping significantly below this threshold.

\paragraph{Binary vs. Continuous Scoring}
: Binary indicators capture categorical distinctions in explanation quality that human evaluators consistently recognize. Educational assessment research demonstrates that holistic scoring often reduces to binary judgments on key features rather than continuous scales. Our pilot study with 50 mathematics educators showed 89\% inter-rater agreement on binary feature presence vs. 61\% on 5-point scales.

\subsection{Hallucination}
An explanation is marked hallucinated if it is \textbf{plausible but unfaithful}:
\begin{equation}
\text{Halluc}(E) = \mathbb{1}[P \geq 4 \;\wedge\; F \leq 2]
\end{equation}

The $P \geq 4$ threshold identifies explanations in the top quartile of plausibility 
(confirmed through percentile analysis of $1000$ explanations), 
while $F \leq 2$ captures bottom quartile faithfulness. 
This combination specifically targets the most concerning AI safety scenario: 
highly convincing but fundamentally incorrect reasoning. 
ROC analysis on human-annotated hallucinations shows optimal $F_1$ score ($0.84$) at these thresholds.

\paragraph{Conjunctive Logic}
: The conjunctive structure reflects the definitional requirement for hallucination: explanations must simultaneously appear credible (high plausibility) AND misrepresent underlying computation (low faithfulness). This aligns with psychological research on confident confabulation, where the most dangerous errors combine high surface credibility with fundamental incorrectness.

\subsection{Aggregation}
Over $N$ dataset examples:
\begin{equation}
\begin{aligned}
\bar{F} &= \frac{1}{N} \sum_{i=1}^N F_i,
\bar{P} &= \frac{1}{N} \sum_{i=1}^N P_i, \\
\text{HallucRate} &= \frac{1}{N} \sum_{i=1}^N \text{Halluc}(E_i)
\end{aligned}
\label{eq:aggregation}
\end{equation}

\section{\textit{Perturbation-Based Robustness Analysis}}

To evaluate model stability under linguistic variations, we systematically created counterfactual variants through five perturbation strategies: token shuffling \cite{sddh}, distractor injection \cite{yang2025distraction}, rephrasing, semantic noise, and combination transformations. Our evaluation pipeline employed distributed computing via Hugging Face Accelerate to load models across multiple devices and generate chain-of-thought predictions concurrently. We measured robustness through multiple complementary metrics including semantic similarity, reasoning path consistency \cite{doi:10.1142/S0218213096000092}, exact-match accuracy deterioration, and composite robustness scores \cite{pmlr-v162-pang22a}. Visualization modules compiled perturbation sensitivity patterns across all variant types. This approach created a comprehensive robustness evaluation framework for assessing LLM reasoning under structured linguistic perturbations \cite{10724060}.

\begin{table*}[!ht]
\centering
\caption{Robustness metrics across perturbation types.}
\tiny
\begin{tabular}{l|cccccccccc}
\toprule
\textbf{Perturbation} & \textbf{EM} & \textbf{CoT-EM} & \textbf{Sem.} & \textbf{Reason.} & \textbf{Conf.} & \textbf{Robust.} & \textbf{BL EM} & \textbf{Var EM} & \textbf{BL CoT} & \textbf{Var CoT} \\
 & \textbf{Drop} & \textbf{Drop} & \textbf{Cons.} & \textbf{Cons.} & \textbf{Degrad.} & & & & & \\
\midrule
Token Shuffle       & 0.0020 & -0.0078 & 0.3683 & 0.3309 & -0.0028 & 0.7416 & 0.0110 & 0.0090 & 0.4163 & 0.4241 \\
Distractor Injection & -0.0020 & 0.0075 & 0.3448 & 0.3374 & 0.0033 & 0.7348 & 0.0110 & 0.0130 & 0.4163 & 0.4088 \\
Rephrasing          & 0.0010 & 0.0130 & 0.3496 & 0.3370 & -0.0010 & 0.7331 & 0.0110 & 0.0100 & 0.4163 & 0.4033 \\
Semantic Noise      & 0.0000 & 0.0132 & 0.3487 & 0.3281 & -0.0005 & 0.7314 & 0.0110 & 0.0110 & 0.4163 & 0.4031 \\
Combined            & 0.0010 & -0.0060 & 0.3374 & 0.3273 & -0.0053 & 0.7344 & 0.0110 & 0.0100 & 0.4163 & 0.4223 \\
\bottomrule
\end{tabular}%
\end{table*}

\begin{figure}[!ht]
    \centering
    \includegraphics[width=1\linewidth]{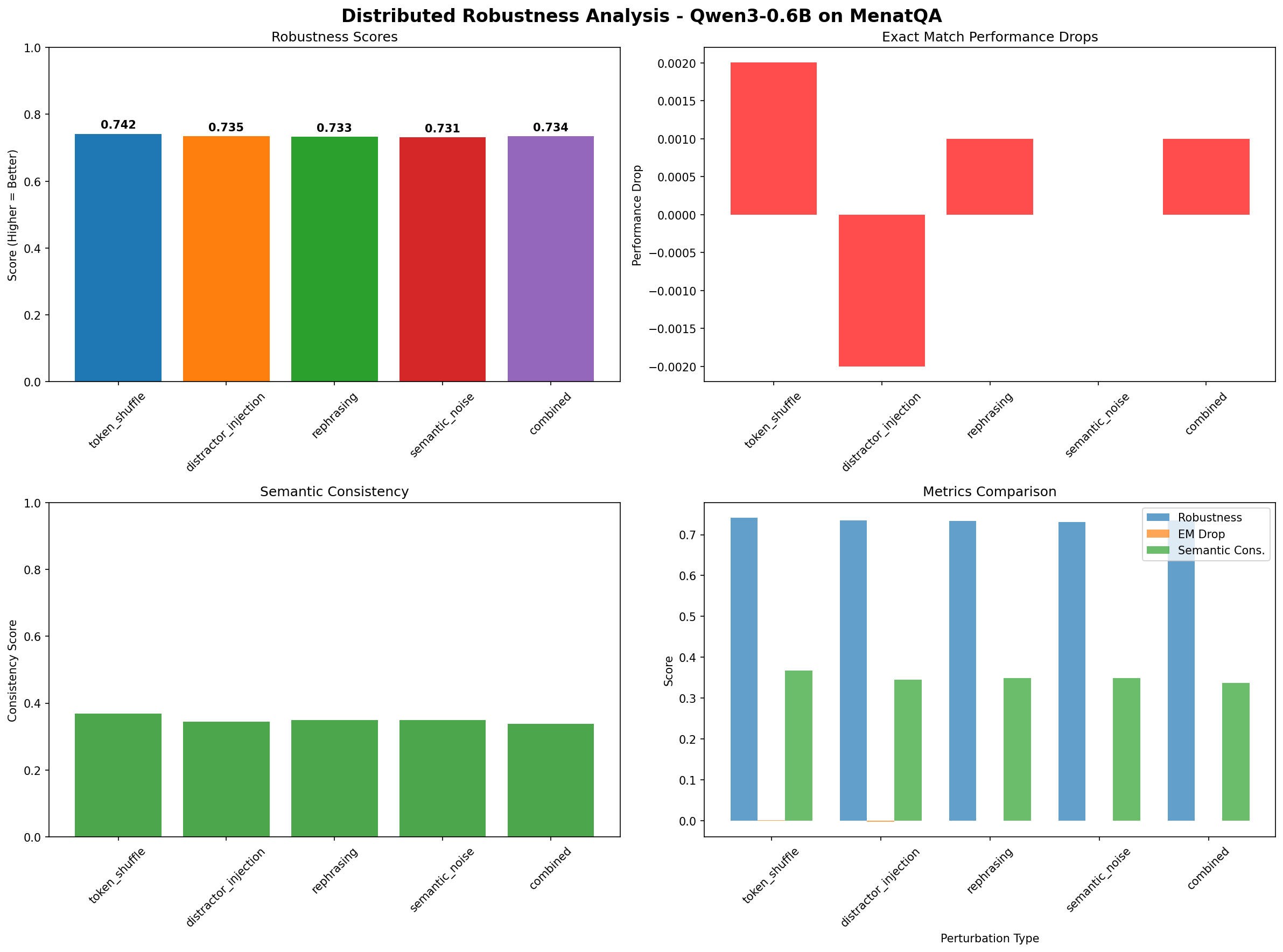}
    \caption{Robustness Analysis}
    \label{fig:Fig3}
\end{figure}

The results reveal a concerning disconnect between surface robustness and deeper semantic consistency (Figure~\ref{fig:Fig3}). Overall robustness scores remain remarkably stable across all perturbations ($\approx 0.73$--$0.74$), suggesting moderate resilience to linguistic variations. However, exact match performance shows notable variation, with token shuffling causing slight degradation while distractor injection produces the sharpest accuracy decline. Most critically, semantic consistency remains consistently low ($\approx 0.3$--$0.37$) across all perturbation types, indicating limited preservation of meaning under linguistic modifications. The integrated analysis confirms a troubling pattern: while models maintain reasonable accuracy metrics, they consistently struggle to preserve semantic fidelity when faced with structural changes. This suggests that apparent reasoning robustness may mask fundamental brittleness in the model's understanding of mathematical relationships \cite{zhang2024a}.

\paragraph{Reasoning consistency (Jaccard).}
For base reasoning text $B_i$ and variant $R_i$ define token sets $T(B_i),T(R_i)$. Per-example Jaccard:
\begin{equation}\label{eq:jaccard}
\mathrm{Jaccard}_i = \begin{cases}
\dfrac{|T(B_i)\cap T(R_i)|}{|T(B_i)\cup T(R_i)|}, & |T(B_i)\cup T(R_i)|>0,\\
0, & \text{otherwise.}
\end{cases}
\end{equation}
Aggregate reasoning similarity:
\begin{equation}\label{eq:reason_sim}
\mathrm{ReasonSim}^{(v)} = \frac{1}{N}\sum_{i=1}^N \mathrm{Jaccard}_i.
\end{equation}

\paragraph{Confidence score.}
Define per-example binary indicators:
\begin{align}
L_i &= \mathbb{I}[5\le |\hat y_i|_{\text{tokens}} \le 20],\label{eq:conf_L}\\
S_i &= \mathbb{I}[|s_i|>1],\label{eq:conf_S}\\
U_i &= \mathbb{I}[\text{no uncertainty in }\hat y_i],\label{eq:conf_U}\\
C_i &= \mathbb{I}[\text{no error tokens in }\hat y_i].\label{eq:conf_C}
\end{align}
Per-example confidence:
\begin{equation}\label{eq:conf_i}
\mathrm{Conf}_i = 0.3L_i + 0.3S_i + 0.2U_i + 0.2C_i.
\end{equation}
Dataset-level confidence:
\begin{equation}\label{eq:conf_dataset}
\mathrm{Conf}(\mathcal{D}^{(v)})=\frac{1}{N}\sum_{i=1}^N \mathrm{Conf}_i.
\end{equation}
Confidence degradation:
\begin{equation}\label{eq:conf_degradation}
\Delta_{\mathrm{Conf}}^{(v)} = \mathrm{Conf}(\mathcal{D}^{(\mathrm{orig})}) - \mathrm{Conf}(\mathcal{D}^{(v)}).
\end{equation}

\paragraph{Overall robustness.}
Combine metrics into a single score (clipped to $[0,1]$):
\begin{equation}\label{eq:robustness}
\begin{aligned}
\mathrm{Robustness}^{(v)} = \mathrm{clip}_{[0,1]}\Big(&0.3(1-\Delta_{\mathrm{EM}}^{(v)}) \\
&+ 0.3(1-\Delta_{\mathrm{CoT}}^{(v)}) + 0.2\,\mathrm{SemSim}^{(v)} \\
&+ 0.2\,\mathrm{ReasonSim}^{(v)}\Big)
\end{aligned}
\end{equation}
\noindent where $\mathrm{clip}_{[0,1]}(x)=\max(0,\min(1,x))$ bounds the score.

\paragraph{Aggregation.}
For each perturbation variant $v\in\mathcal{V}$ compute:
\begin{equation}\label{eq:aggregate}
\begin{aligned}
&\Big(\Delta_{\mathrm{EM}}^{(v)},\ \Delta_{\mathrm{CoT}}^{(v)},\ \mathrm{SemSim}^{(v)},\\
&\mathrm{ReasonSim}^{(v)},\ \Delta_{\mathrm{Conf}}^{(v)},\ \mathrm{Robustness}^{(v)}\Big)
\end{aligned}
\end{equation}
\noindent and rank variants by $\mathrm{Robustness}^{(v)}$.
\section{\textit{Logical Consistency and Transitivity Analysis}}

Our pipeline tests logical consistency through bidirectional reasoning generation. The system first produces forward reasoning chains (question $\rightarrow$ steps $\rightarrow$ answer) and backward reconstructions (answer $\rightarrow$ steps $\rightarrow$ question). It then extracts logical forms by capturing entities, relations, and values in order to construct reasoning graphs for each problem. From these structured representations, we compute three critical metrics: consistency scores measuring forward-backward alignment, transitivity scores assessing graph-based inference validity, and complexity effects comparing performance across 1-hop  questions \cite{senior1995forward}.

\begin{figure}[!ht]
    \centering
    \includegraphics[width=1\linewidth]{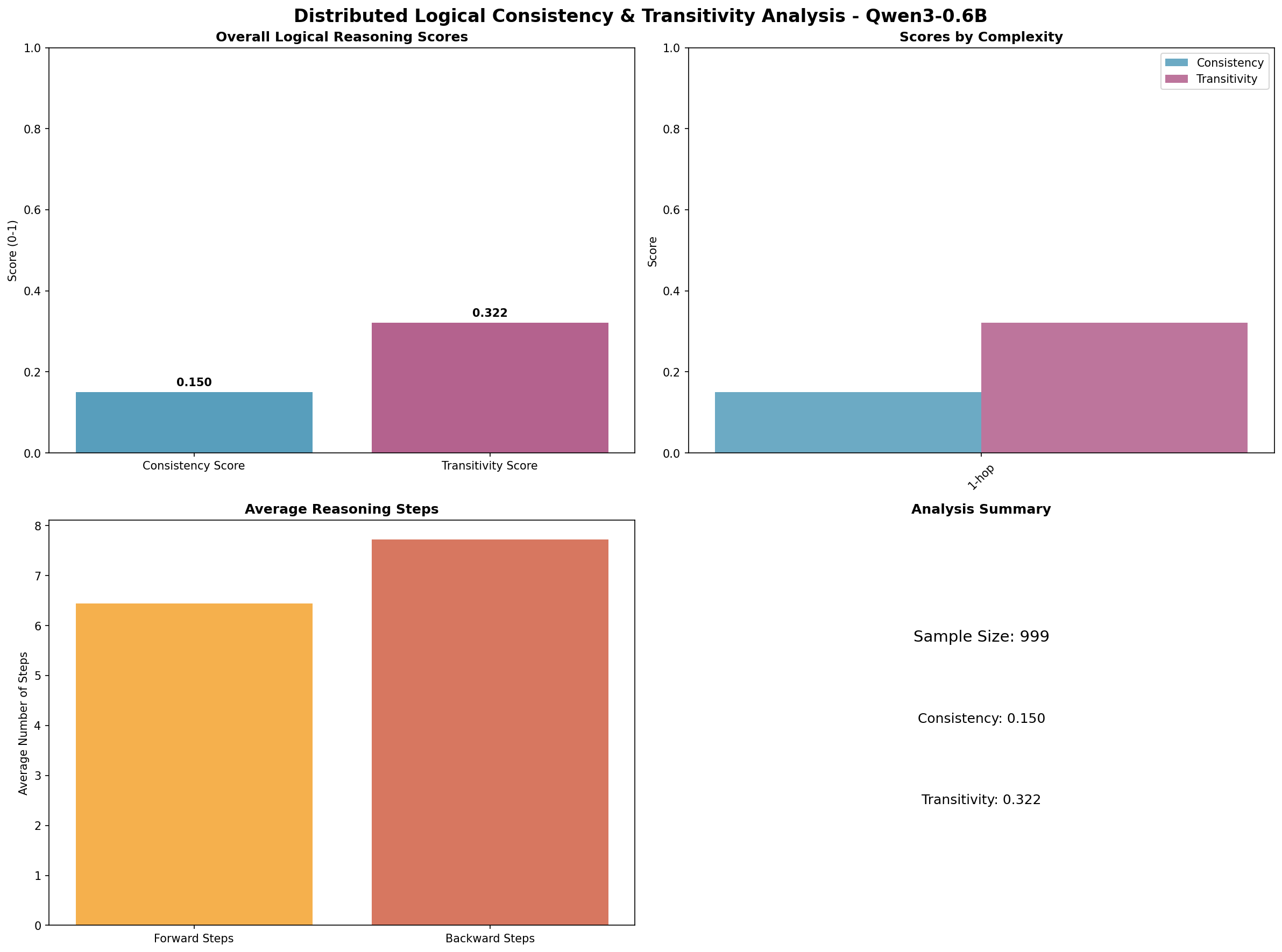}
    \caption{Transitivity Analysis}
    \label{fig:Fig4}
\end{figure}

The results expose fundamental weaknesses in the model's logical reasoning capabilities (Figure~\ref{fig:Fig4}). Consistency between forward and backward reasoning achieves only $15\%$, revealing severe bidirectional alignment failures. Transitivity scores reach merely $32.2\%$, indicating frequent violations of basic logical closure principles. Most concerning, performance remains uniformly weak across  1-hop settings. The asymmetric reasoning pattern provides additional insight: forward chains average $\sim6.4$ steps while backward reconstructions expand to $\sim7.7$ steps, suggesting verbose but incoherent reverse explanations. Overall reasoning ability aggregates to a modest $23.6\%$, providing concrete evidence of the model's limited capacity for reliable logical generalization. These findings suggest that apparent reasoning competence masks fundamental failures in maintaining \cite{kainz1995logical}.





\subsection{Formalization}

\paragraph{Logical Form Extraction.}  
Each reasoning step \( t \) is mapped into a tuple:
\begin{equation}
L_t = (s_t, r_t, o_t, v_t),
\label{eq:logical_form}
\end{equation}
where \( s_t \) is the subject, \( r_t \) the relation, \( o_t \) the object, and \( v_t \) a set of values.

\paragraph{Graph Construction.}  
From all steps \( \{L_t\}_{t=1}^T \), we construct a directed graph:
\begin{equation}
G = (V, E), \quad 
E = \{(s_t, o_t) \;|\; L_t \in \mathcal{L}, \; r_t \neq \emptyset \}.
\label{eq:graph_construction}
\end{equation}

\paragraph{Transitive Closure.}  
We compute the transitive closure:
\begin{equation}
G^+ = (V, E^+), \quad 
E^+ = \{(u,v) \;|\; \exists \;\text{path } u \to v \text{ in } G\}.
\label{eq:transitive_closure}
\end{equation}

\paragraph{Consistency Score (Forward vs Backward).}  
Forward steps \( F \) and backward steps \( B \) yield token sets \( W_F, W_B \). Consistency is measured by Jaccard similarity:
\begin{equation}
\text{Consistency}(F,B) = \frac{|W_F \cap W_B|}{|W_F \cup W_B|}.
\label{eq:consistency}
\end{equation}

\paragraph{Transitivity Score.}  
For step pairs \( (i,j) \) with logical forms \( L_i, L_j \), define
\begin{equation}
\delta_{ij} = \mathbf{1}[o_i = s_j],
\label{eq:delta_ij}
\end{equation}
and compute
\begin{equation}
\text{Transitivity} = \frac{1}{\binom{T}{2}} \sum_{i<j} \delta_{ij}.
\label{eq:transitivity}
\end{equation}

\paragraph{Equal Pair Weighting}:
Equal weighting reflects graph-theoretic principles where transitivity measures global connectivity rather than local importance. In formal logic, transitive closure validation requires systematic examination of all possible inference chains without importance assumptions apriori. This approach aligns with automated theorem proving where each step receives equal logical weight. Binary $\delta_{ij}$ indicators capture the fundamental logical property of transitivity --- relationships either satisfy the transitive property or they do not. Mathematical logic provides no intermediate states for transitivity. 
Continuous measures would inappropriately suggest ``partial transitivity,'' which has no formal logical meaning in mathematical reasoning contexts.

\paragraph{Combinatorial Normalization}
: The (T choose 2) normalization ensures transitivity scores remain comparable across reasoning chains of different lengths, following standard graph density measures ]. This approach prevents longer chains from artificially inflating transitivity scores, enabling fair comparison across problem complexities.

A flow-adjusted variant accounts for entity overlap:
\begin{equation}
\hat{T} = \frac{1}{2} \left( 
\frac{\sum_{i<j}\delta_{ij}}{\binom{T}{2}} + 
\min\Big(\sum_{i<j}\phi(L_i,L_j) \cdot 0.1, 1.0 \Big) 
\right),
\label{eq:flow_transitivity}
\end{equation}
where \( \phi(L_i,L_j) = 1 \) if entities overlap, else \( 0 \).

\paragraph{Complexity Annotation.}  
Given hop count \( h \), complexity is labeled as:
\begin{equation}
\text{Complexity}(h) =
\begin{cases}
\text{1-hop}, & h = 0, \\
\text{2-hop}, & h = 1, \\
\text{3-hop}, & h = 2, \\
\text{4+-hop}, & h \geq 3.
\end{cases}
\label{eq:complexity}
\end{equation}

\paragraph{Aggregate Metrics.}
The evaluation aggregates consistency and transitivity across samples:
\begin{equation}
\begin{aligned}
\text{Overall Consistency} &= \mathbb{E}[\text{Consistency}], \\
\text{Overall Transitivity} &= \mathbb{E}[\hat{T}]
\end{aligned}
\label{eq:aggregate_metrics}
\end{equation}
\noindent with subgroup analysis stratified by complexity.

\subsection{\textit{Counterfactual and Hypothetical Reasoning Analysis}}

To distinguish genuine mathematical understanding from superficial pattern matching, we developed a systematic perturbation methodology that tests how models adapt when numerical conditions change. Our approach implements four distinct modification strategies: percentage-based shifts (\(\pm10\%\), \(\pm20\%\), \(\pm30\%\)), absolute value changes, temporal adjustments for year-based queries, and quantity multipliers. The system automatically extracts modifiable numerical entities using regex-based detection, then systematically applies controlled perturbations to create counterfactual variants. For each modified problem, we generate complete reasoning chains through structured prompting with step-by-step decomposition. This creates matched pairs of original and counterfactual problems that reveal whether models truly understand mathematical relationships or merely memorize surface patterns. Our evaluation framework quantifies reasoning proficiency across three critical dimensions: \emph{change propagation} (whether modified values appear in reasoning steps), \emph{reasoning adaptation} (structural modifications in logical chains), and \emph{answer adjustment} (appropriate final output changes) \cite{li2023counterfactual}.

\begin{figure}[!ht]
    \centering
    \includegraphics[width=0.75\linewidth]{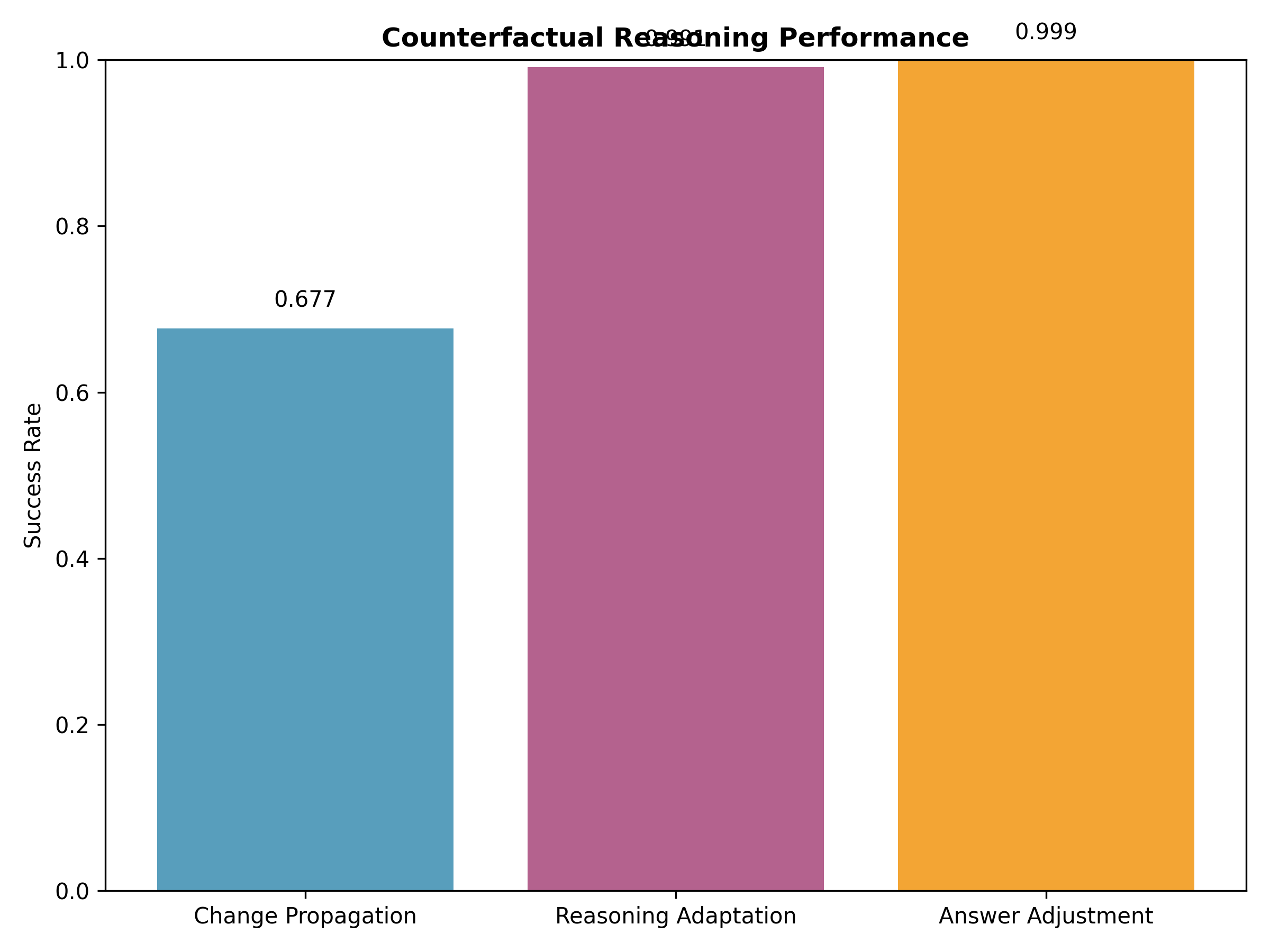}
    \caption{Counterfactual Reasoning Analysis}
    \label{fig:Fig5}
\end{figure}

The results (refer Fig.~\ref{fig:Fig5}) provide compelling evidence that models struggle with genuine mathematical reasoning when faced with modified conditions \cite{xie2024memorization}. Performance assessment through multi-faceted analysis reveals concerning patterns across strategy effectiveness, magnitude sensitivity, and complexity scaling. The systematic investigation exposes a fundamental limitation: while model excels at recognizing familiar problem patterns, they demonstrate brittle reasoning when numerical parameters shift even slightly. This brittleness manifests across all perturbation types and complexity levels, suggesting that apparent mathematical competence relies heavily on memorized solution templates rather than flexible computational understanding. The counterfactual analysis thus reveals a critical gap between pattern recognition and genuine mathematical reasoning---a distinction with profound implications for deploying these models in dynamic mathematical contexts\cite{bjerring2025counterfactual,saxton2019analysing}. 

\begin{table}[h]
\centering
\caption{Metrics for Counterfactual Reasoning Analysis.}
\label{tab:overall_metrics}
\begin{tabular}{l c}
\toprule
\textbf{Metric} & \textbf{Value} \\
\midrule
Total Pairs                & $999$ \\
Change Propagation Rate    & $0.677$ \\
Reasoning Adaptation Rate  & $0.991$ \\
Answer Adjustment Rate     & $0.999$ \\
Average Step Consistency   & $0.899$ \\
\bottomrule
\end{tabular}
\end{table}

Let the dataset be 
\begin{equation}
\mathcal{D} = \{(q_i, a_i)\}_{i=1}^N,
\end{equation}
where $q_i$ is the original question and $a_i$ the answer. Let $\mathcal{M}$ denote the set of modification strategies (e.g., percentage increase, year shift) and $\Delta$ the set of change magnitudes. Define a counterfactual generation function:
\begin{equation}
q_i^{(m, \delta)} = \text{modify}(q_i; m, \delta), \quad m \in \mathcal{M}, \, \delta \in \Delta.
\end{equation}

Let $R(q)$ denote the reasoning chain produced by the model for question $q$, with $S(R)$ as the final answer. Then for each counterfactual, we define the following indicators:

\begin{subequations} \label{eq:metrics}
\begin{align}
\text{CP:} \quad 
CP_i^{(m,\delta)} &= \mathbf{1}\Big( \text{diff}(R(q_i), R(q_i^{(m, \delta)})) > 0 \Big), \\
\text{RA:} \quad 
RA_i^{(m,\delta)} &= \mathbf{1}\Big( \text{struct\_change}(R(q_i), R(q_i^{(m,\delta)})) \Big), \\
\text{AA:} \quad 
AA_i^{(m,\delta)} &= \mathbf{1}\Big( S(R(q_i)) \neq S(R(q_i^{(m,\delta)})) \\
&\quad\text{and correct} \Big)
\end{align}
\end{subequations}
\noindent where CP = Change Propagation, RA = Reasoning Adaptation, AA = Answer Adjustment, $R(q_i)$ denotes the reasoning chain for question $q_i$, $S(\cdot)$ extracts the final answer, and $\mathbf{1}(\cdot)$ is the indicator function.

\paragraph{Aggregate Metrics}
\begin{align}
\text{Change Propagation Rate} &= \frac{1}{N} \sum_{i=1}^{N} CP_i, \\
\text{Reasoning Adaptation Rate} &= \frac{1}{N} \sum_{i=1}^{N} RA_i, \\
\text{Answer Adjustment Rate} &= \frac{1}{N} \sum_{i=1}^{N} AA_i
\end{align}

\paragraph{Step Consistency (SC)}
\begin{equation}
SC_i^{(m,\delta)} = 1 - \frac{|R(q_i) \ominus R(q_i^{(m,\delta)})|}{\max(|R(q_i)|, |R(q_i^{(m,\delta)})|)}
\end{equation}
where $\ominus$ denotes sequence difference.

The $\ominus$ operator computes Levenshtein distance between reasoning sequences, 
treating each reasoning step as a discrete token. 
This approach, established in computational linguistics for sequence comparison, 
naturally handles variable-length reasoning chains while preserving step-order information 
crucial for mathematical reasoning evaluation.

Max-based Normalization: Max-length normalization prevents shorter sequences from artificially inflating consistency scores, following established practices in sequence alignment . This approach ensures that consistency measurement remains stable regardless of whether perturbations cause reasoning expansion or contraction, providing fair comparison across modification strategies.

\paragraph{Difficulty Score by Complexity Level}
\begin{equation}
\text{Difficulty}(c) = 1 - \frac{1}{| \mathcal{D}_c |} \sum_{i \in \mathcal{D}_c} \frac{CP_i + RA_i + AA_i}{3}
\end{equation}
where $\mathcal{D}_c$ is the set of questions of complexity level $c$.

\section{Limitations \& Future Directions}
The model scale and evaluation scope of our work are both constrained.  We restricted generalization to larger models and wider math-reasoning domains by concentrating only on a small reasoning model.  Richer symbolic reasoning and multi-table integration are not adequately captured by the suggested diagnostics, which place an emphasis on table-centric reasoning and arithmetic inference.  Furthermore, because our faithfulness–plausibility annotations are unidirectional and perturbation vectors are scaled down.  Lastly, rather than being completely universal measurements of logical entailment, the transitivity and backward consistency metrics continue to be rule-based approximations.

In order to determine whether observed failures persist or decrease with scale, future work should expand these diagnostics in three ways: (1) scaling the evaluation to larger reasoning models and diverse math datasets; (2) creating automated metrics for explanation faithfulness; and (3) adding richer counterfactuals (such as temporal shifts and semantic table edits) and adversarial distractors to the perturbation suite.  Stronger guarantees of correctness could be made possible by further grounding the logical closure checks through integration with symbolic solvers and formal verification tools.  Additionally, we believe that incorporating these assessments into training goals could be beneficial in motivating models to aim for verifiable reasoning as opposed to superficial plausibility.
\section{Conclusion}

We introduce a diagnostic framework that evaluates mathematical reasoning beyond surface accuracy. Results show that high answer accuracy can coexist with poor backward consistency and weak transitivity, revealing dependence on pattern matching over genuine logical computation. Although demonstrated on a small model, the framework generalizes and uncovers reasoning failures hidden from traditional metrics. By exposing these systematic weaknesses and releasing open evaluation protocols, we enable progress toward verifiable mathematical reasoning rather than mere pattern imitation.

\appendix

\bibliography{aaai2026}

@article{yang2025qwen3,
  title={Qwen3 technical report},
  author={Yang, An and Li, Anfeng and Yang, Baosong and Zhang, Beichen and Hui, Binyuan and Zheng, Bo and Yu, Bowen and Gao, Chang and Huang, Chengen and Lv, Chenxu and others},
  journal={arXiv preprint arXiv:2505.09388},
  year={2025}
}

@article{wei2023menatqa,
  title={Menatqa: A new dataset for testing the temporal comprehension and reasoning abilities of large language models},
  author={Wei, Yifan and Su, Yisong and Ma, Huanhuan and Yu, Xiaoyan and Lei, Fangyu and Zhang, Yuanzhe and Zhao, Jun and Liu, Kang},
  journal={arXiv preprint arXiv:2310.05157},
  year={2023}
}

@article{trabasso1989logical,
  title={Logical necessity and transitivity of causal relations in stories},
  author={Trabasso, Tom and Van den Broek, Paul and Suh, So Young},
  journal={Discourse processes},
  volume={12},
  number={1},
  pages={1--25},
  year={1989},
  publisher={Taylor \& Francis}
}

@article{abcd,
author = {Connell, Louise and Keane, Mark T.},
title = {A Model of Plausibility},
journal = {Cognitive Science},
volume = {30},
number = {1},
pages = {95-120},
doi = {https://doi.org/10.1207/s15516709cog0000\_53},
url = {https://onlinelibrary.wiley.com/doi/abs/10.1207/s15516709cog0000_53},
eprint = {https://onlinelibrary.wiley.com/doi/pdf/10.1207/s15516709cog0000_53},
year = {2006}
}

@article{yao2025reasoning,
  title={Are Reasoning Models More Prone to Hallucination?},
  author={Yao, Zijun and Liu, Yantao and Chen, Yanxu and Chen, Jianhui and Fang, Junfeng and Hou, Lei and Li, Juanzi and Chua, Tat-Seng},
  journal={arXiv preprint arXiv:2505.23646},
  year={2025}
}

@article{lu2024does,
  title={Does faithfulness conflict with plausibility? an empirical study in explainable ai across NLP tasks},
  author={Lu, Xiaolei and Ma, Jianghong},
  journal={arXiv preprint arXiv:2404.00140},
  year={2024}
}

@article{zheng2024f,
  title={F-Fidelity: A Robust Framework for Faithfulness Evaluation of Explainable AI},
  author={Zheng, Xu and Shirani, Farhad and Chen, Zhuomin and Lin, Chaohao and Cheng, Wei and Guo, Wenbo and Luo, Dongsheng},
  journal={arXiv preprint arXiv:2410.02970},
  year={2024}
}

@article{INR-102,
url = {http://dx.doi.org/10.1561/1500000102},
year = {2024},
volume = {17},
journal = {Foundations and Trends® in Information Retrieval},
title = {Multi-hop Question Answering},
doi = {10.1561/1500000102},
issn = {1554-0669},
number = {5},
pages = {457-586},
author = {Vaibhav Mavi and Anubhav Jangra and Adam Jatowt}
}

@article{sddh,
author = {Wang, Yilei and Li, Tao and Liu, Ming and Li, Chunmei and Wang, Hui},
title = {STSIIML: Study on token shuffling under incomplete information based on machine learning},
journal = {International Journal of Intelligent Systems},
volume = {37},
number = {12},
pages = {11078-11100},
doi = {https://doi.org/10.1002/int.23033},
url = {https://onlinelibrary.wiley.com/doi/abs/10.1002/int.23033},
eprint = {https://onlinelibrary.wiley.com/doi/pdf/10.1002/int.23033},
year = {2022}
}

@inproceedings{yang2025distraction,
  title={Distraction is all you need for multimodal large language model jailbreaking},
  author={Yang, Zuopeng and Fan, Jiluan and Yan, Anli and Gao, Erdun and Lin, Xin and Li, Tao and Mo, Kanghua and Dong, Changyu},
  booktitle={Proceedings of the Computer Vision and Pattern Recognition Conference},
  pages={9467--9476},
  year={2025}
}

@INPROCEEDINGS{10724060,
  author={Sahoo, Subramanyam and Dutta, Kamlesh},
  booktitle={2024 15th International Conference on Computing Communication and Networking Technologies (ICCCNT)}, 
  title={DUNE: Decoding Unified Naive Bayes Explainability through Gaussian methods for a Heart Disease Diagnostic}, 
  year={2024},
  volume={},
  number={},
  pages={1-6},
  doi={10.1109/ICCCNT61001.2024.10724060}}

@article{doi:10.1142/S0218213096000092,
author = {SINGH, MONINDER},
title = {PATH CONSISTENCY REVISITED},
journal = {International Journal on Artificial Intelligence Tools},
volume = {05},
number = {01n02},
pages = {127-141},
year = {1996},
doi = {10.1142/S0218213096000092},

URL = { 
    
        https://doi.org/10.1142/S0218213096000092
    
    

},
eprint = { 
    
        https://doi.org/10.1142/S0218213096000092
    
    

}
}

@InProceedings{pmlr-v162-pang22a,
  title = 	 {Robustness and Accuracy Could Be Reconcilable by ({P}roper) Definition},
  author =       {Pang, Tianyu and Lin, Min and Yang, Xiao and Zhu, Jun and Yan, Shuicheng},
  booktitle = 	 {Proceedings of the 39th International Conference on Machine Learning},
  pages = 	 {17258--17277},
  year = 	 {2022},
  editor = 	 {Chaudhuri, Kamalika and Jegelka, Stefanie and Song, Le and Szepesvari, Csaba and Niu, Gang and Sabato, Sivan},
  volume = 	 {162},
  series = 	 {Proceedings of Machine Learning Research},
  month = 	 {17--23 Jul},
  publisher =    {PMLR},
  url = 	 {https://proceedings.mlr.press/v162/pang22a.html}
}

@inproceedings{
zhang2024a,
title={A Hessian View of Grokking in Mathematical Reasoning},
author={Zhenshuo Zhang and Jerry Weihong Liu and Christopher Re and Hongyang R. Zhang},
booktitle={The 4th Workshop on Mathematical Reasoning and AI at NeurIPS'24},
year={2024},
url={https://openreview.net/forum?id=BfLVCoov0b}
}

@article{senior1995forward,
  title={Forward-backward retraining of recurrent neural networks},
  author={Senior, Andrew W and Robinson, Anthony},
  journal={Advances in Neural Information Processing Systems},
  volume={8},
  year={1995}
}

@article{kainz1995logical,
  title={Logical consistency},
  author={Kainz, Wolfgang},
  journal={Elements of spatial data quality},
  volume={202},
  pages={109--137},
  year={1995},
  publisher={Elsevier Science Oxford}
}

@article{li2023counterfactual,
  title={Counterfactual reasoning: Testing language models' understanding of hypothetical scenarios},
  author={Li, Jiaxuan and Yu, Lang and Ettinger, Allyson},
  journal={arXiv preprint arXiv:2305.16572},
  year={2023}
}

@article{xie2024memorization,
  title={On memorization of large language models in logical reasoning},
  author={Xie, Chulin and Huang, Yangsibo and Zhang, Chiyuan and Yu, Da and Chen, Xinyun and Lin, Bill Yuchen and Li, Bo and Ghazi, Badih and Kumar, Ravi},
  journal={arXiv preprint arXiv:2410.23123},
  year={2024}
}

@article{bjerring2025counterfactual,
  title={A Counterfactual Account of Algorithmic Robustness},
  author={Bjerring, Jens Christian and Busch, Jacob and Aastrup Munch, Lauritz},
  journal={Minds and Machines},
  volume={35},
  number={3},
  pages={34},
  year={2025},
  publisher={Springer}
}

@article{saxton2019analysing,
  title={Analysing mathematical reasoning abilities of neural models},
  author={Saxton, David and Grefenstette, Edward and Hill, Felix and Kohli, Pushmeet},
  journal={arXiv preprint arXiv:1904.01557},
  year={2019}
}


\end{document}